\documentclass{article}

\usepackage{arxiv}

\usepackage[utf8]{inputenc} 
\usepackage[T1]{fontenc}    
\usepackage{hyperref}       
\usepackage{url}            
\usepackage{booktabs}       
\usepackage{amsfonts}       
\usepackage{nicefrac}       
\usepackage{microtype}      
\usepackage{lipsum}
\usepackage{graphicx}
\graphicspath{ {./images/} }

\title{Sentiment Analysis Using Averaged Weighted Word Vector Features}

\author{
 Ali Erkan \\
  Computer Engineering Department\\
  Boğaziçi University\\
  İstanbul, Turkey \\
  \texttt{ali.erkan@boun.edu.tr} \\
   \And
 Tunga Güngör\\
  Computer Engineering Department\\
  Boğaziçi University\\
  İstanbul, Turkey \\
  \texttt{gungort@boun.edu.tr} \\
}

\usepackage{pgfplots}
\pgfplotsset{compat=newest}
\pgfplotsset{
	tick label style = {font = \normalsize},
	label style = {font = \normalsize},
	legend style = {font = \normalsize},
	every axis plot/.append style = {font = \normalsize},
	xticklabel style = {font=\normalsize}
}
\usetikzlibrary{matrix,chains,positioning,decorations.pathreplacing,arrows}

\begin{document}
\maketitle
\begin{abstract}
People use the World Wide Web heavily to share their experiences with entities such as products, services, or travel destinations. Texts that provide online feedback in the form of reviews and comments are essential to make consumer decisions. These comments create a valuable source that may be used to measure satisfaction related to products or services. Sentiment analysis is the task of identifying opinions expressed in such text fragments. In this work, we develop two methods that combine different types of word vectors to learn and estimate the polarity of reviews. We develop average review vectors from word vectors and add weights to these
review vectors using word frequencies in positive and negative sensitivity-tagged reviews. We applied the methods to several datasets from diﬀerent domains used as standard sentiment analysis benchmarks. We ensemble the techniques with each other and existing methods, and we make a comparison with the approaches in the literature. The results show that the performances of our approaches outperform the state-of-the-art success rates.
\end{abstract}

\section*{Introduction}
Sentiment analysis is the process of computationally identifying and categorizing opinions expressed in a text, especially in order to determine whether the writer's attitude toward a particular topic, product, etc., is positive, negative, or neutral. A basic task in sentiment analysis is classifying the polarity of a given text in the document or sentence.

As mentioned in Task 5 of Semeval 2016\cite{Semeval5:16}, we use the Web to share our experiences about products, services, or travel destinations \cite{Yoo:08}. Texts that provide online feedback, such as reviews, comments, etc., are important for consumer decision-making \cite{Dina:06} and customer comments create valuable sources that can help companies measure satisfaction and improve their products or services.

Sentiment Analysis (SA) includes every aspect of Natural Language Processing such as entity recognition, coreference resolution, negation handling, etc. \cite{Liu:12} and as Cambria et al.\cite{Cambria:13} mention, “it requires a deep understanding of the explicit and implicit, regular and irregular, and syntactic and semantic language rules”. The several workshops and conferences focusing on SA-related shared tasks. Some of them are (NTCIR\cite{Seki:10}; TAC2013\cite{Mitchell:13}; SemEval-2013 Task 2\cite{Nakov:13}; SemEval-2014 Task 4\cite{Rosenthal:14};  SemEval-2015 Task 12\cite{Pontiki:15}; SemEval-2016 Task 6\cite{Muhammad:16}; SemWebEval 2014\cite{Recupero:14}; GESTALT-2014\cite{Ruppenhofer:14}; SentiRuEval\cite{Loukachevitch:15}. These competitions provide training datasets. Currently, most of the available SA-related datasets (Socher et al\cite{Socher:13}; Ganu et al\cite{Ganu:09}) are monolingual and generally focus on English texts. Also, some of them provide multilingual datasets (Klinger and Cimiano \cite{Klinger:14}; Jiménez-Zafra et al.\cite{Jimenez:15}) that are useful to enable the development and testing of cross-lingual methods \cite{Lambert:15}.


Besides these, some other datasets exist, such as Stanford IMDB Reviews\cite{Stanford:11} or Yelp dataset \cite{Yelp:18}, and some SA studies were made by using these datasets. In this work, we focus on a comprehensive analysis of SA studies, and we are very close to the state-of-the-art results (\cite{Mesnil:15},  \cite{Wang:12}). 

In this work, we make a comprehensive analysis of some existing methods to produce semantic polarities from the reviews from different domains, and we propose two new approaches to produce semantic polarities. Our approach uses weighted word vectors to create feature sets. For this purpose, we used word2vec\cite{Mikolov:13} and glove\cite{pennington:14} methods. We ensembled and compared our approaches with existing approaches and analyzed experimental results from domains IMBD reviews, Semeval 2016 Task Dataset, and Yelp Restaurant reviews. We are very close to state-of-the-art performance accuracies with our approaches.

We provide the source code and prepared datasets for the model as well as trained word vectors at https://github.com/alierkan/Sentiment-Analysis.

The rest of this paper is organized as follows: Section 2 presents previous works related to sentiment analysis. In Section 3, we describe our model. In Section 4, we describe the data sets used during experimentation. The results are then presented in Section 5. Finally, Section 6 summarizes the final conclusions and potential future lines of this work.

\section*{Related Works}
In the literature, there are some datasets to use models, such as Stanford IMDB Movie Reviews, SEMEVAL restaurant and laptop reviews, and the Yelp dataset. We mentioned previous studies that used these datasets. Although there are some rule-based studies in the literature, we listed only studies that used learning algorithms since our study is also focused on machine learning algorithms.  

Wang and Manning\cite{Wang:12} used Multinomial Naive Bayes (MNB), support vector machine (SVM), and SVM with NB (NBSVM) feature to find out the polarities of the reviews. They split at spaces for unigrams and filtered out anything that is not [A-Za-z] for bigrams. Their approach computes a log-ratio vector between the average word counts extracted from positive documents and the average word counts extracted from negative documents. NBSVM obtained $91.2 \%$ accuracy for the IMDB dataset.

Mesnil et al.\cite{Mesnil:15} used three approaches to discriminate positive and negative sentiment for the IMDB reviews, and then they combined these approaches to reach better accuracy. Their first approach is computing the probability of the test document belonging to the positive and negative class via Bayes’ rule by using n-grams and Recurrent Neural Networks (RNNs) \cite{Mikolov:10}. As a second approach, they used a supervised reweighing of the counts as in the Naive Bayes Support Vector Machine (NB-SVM) that was mentioned in the previous paragraph\cite{Wang:12}. Finally, they used a sentence vector method, which is proposed by \cite{Le:14} to learn distributed representations of words and paragraphs. The sentence vector was created by using the word2vec algorithm, which was proposed by \cite{Mikolov:13}. To create review vectors, at the first step, they added a unique ID at the beginning of every review. In this way, this id became a word that represents the review. Then, they ran the word2vec algorithm on these modified reviews and prevented the algorithm from removing review IDs. So, they had a matrix that included word vectors of the words in the reviews and review IDs. Every row represents a word vector of the corresponding word. Then, they created a sub-matrix of this matrix that contains only word vectors of review IDs. This method has one major issue: A review vector should be generated from both training and test reviews. So, if you want to find the polarity of a new review, all steps, including creating review vectors, have to be repeated. Therefore it is not a practical method. Then, they combined the results of the three approaches and achieved higher accuracy. Mesnil et al\cite{Mesnil:15} passed Wang and Manning\cite{Wang:12} score and they reached $92.57\%$ accuracy.

In Semeval 2016 Task 5 Subtask 2\cite{Semeval5:16}, a set of customer reviews about a target entity (e.g. a laptop or a restaurant) was given; the goal is to identify a set of {aspect, polarity} tuples that summarize the opinions expressed in each review. Khalil et al.(NileTMRG Team)\cite{Khalil:16} incorporated domain and aspect information in one ensemble classifier consisting of three CNNs\cite{Kim:14} trained using the whole training data provided in both domains and initialized with word vectors that were fine-tuned using training examples collected in a semi-supervised way by the same CNN architecture. Each one of the three classifiers is similar to the one with a slight variation resulting from incorporating domain and aspect knowledge into the CNN model. This incorporation was done by introducing new binary features to the hidden layer of the CNN. The new features indicate the presence or the absence of a certain aspect or domain in a given sentence. They have mostly employed Static-CNN, where initialized input vectors are kept as is, and Dynamic-CNN, where input vectors are updated for optimizing the network. Dynamic-CNN on sentences tokenized from the Yelp academic dataset reviews in restaurants. They ensembled their results. This ensemble model counts votes from three classifiers and predicts the class that has the maximum number of votes from the three classes, namely the positive, the negative, and the neutral. They obtained 85.448 percent accuracy for the English Restaurant dataset of Semeval 2016 Task 5\cite{Semeval5:16}.

Kumar et al.(IIT-TUDA team)\cite{Kumar:16} used Lexical Acquisition and supervised classification using Support Vector Machine SVM at Semeval 2016 Task 5. They used lexical expansion to induce sentiment words based on the distributional hypothesis. Due to their observation of rare words, unseen instances, and limited coverage of available lexicons, they thought that the distributional expansion might be a useful back-off technique (Govind et al. \cite{Govind:14}). They constructed a polarity lexicon for all languages using an external corpus and seed sentiment lexicon. Finally, they computed the normalized positive, negative, and neutral scores for each word similar to (Kumar et al. \cite{Kumar:15}). Their main assumption is that words with the same sentiment are semantically more similar. Hence, words that appear more in positive (negative/neutral) reviews have a higher positive (negative/neutral) sentiment score. They obtained 86.729 percent accuracy for the English Restaurant dataset of Semeval 2016 Task 5\cite{Semeval5:16}.

Brun et al. (XCRE team)\cite{Brun:16} apply a term-centric method for feature extraction. For a term, the features are obtained as the lexical-semantic categories (food, service, etc.) associated with the term by a semantic parser, bigrams and trigrams involving the term, and all syntactic dependencies (subject, object, modifier, attribute, etc.) involving the term. First, aspects are extracted using a conditional random field (CRF) model. Then, aspect categories are found and added as features for polarity classification. The features are also delexicalized, replacing a term with its generic aspect category (e.g. “staff” is replaced by “service”, “sushi” is replaced by “food”). They obtained 88.13\% accuracy for the English restaurant dataset of the Semeval 2016 Task 5\cite{Semeval5:16}.

Jiang et al. (ECNU team)\cite{Jiang:16} employ the Logistic Regression algorithm with the default parameter implemented in lib-linear tools\footnote{https://www.csie.ntu.edu.tw/cjlin/liblinear/} to build the classifiers. The 5-fold cross-validation is adopted for system development. They used linguistic features (Word N-grams, Lemmatized Word N-grams, POS, etc.), sentiment lexicon features (mainly ratios between positive, negative, and potential words related to a given aspect), topic model (the document distribution among predefined topics, the topic probability of each word indicates its significance in corresponding topic) and word2vec features to learn. 

\section*{Method}
By using machine learning techniques to learn the polarity of a review, we should represent it with some features. As mentioned before, they may be a bag of words, the number of words in a sentence, or other features. Nowadays, most of these types of studies include word vectors. 

Word vector models (WVM) represent words in a continuous vector space where semantically similar words are mapped to nearby points. WVM's have a long, rich history in NLP, but all methods depend in some way or another on the Distributional Hypothesis, which states that words that appear in the same contexts share semantic meaning. The different approaches that leverage this principle can be divided into two categories: count-based methods (e.g. Latent Semantic Analysis), and predictive methods (e.g. neural probabilistic language models). Count-based methods compute the statistics of how often some word co-occurs with its neighbor words in a large text corpus and then map these count statistics down to a small, dense vector for each word. Predictive models directly try to predict a word from its neighbors in terms of learned small, dense embedding vectors (considered parameters of the model). Two featured word vector algorithms exist; one is Word2vec, and another is Glove.

Word2vec is a group of related models that are used to produce word embeddings. These models are shallow, two-layer neural networks that are trained to reconstruct linguistic contexts of words. Word2vec takes as its input a large corpus of text and produces a vector space, typically of several hundred dimensions, with each unique word in the corpus being assigned a corresponding vector in the space. Word vectors are positioned in the vector space such that words that share common contexts in the corpus are located in close proximity to one another in the space\cite{Mikolov:13}. We used the Word2vec algorithm with the Continuous Skip-gram Model. It tries to maximize the classification of a word based on another word in the same sentence. More precisely, each current word is used as an input to a log-linear classifier with a continuous projection layer and predicts words within a certain range before and after the current word. Increasing the range improves the quality of the resulting word vectors, but it also increases the computational complexity.  

Glove is an unsupervised learning algorithm for obtaining vector representations for words. Training is performed on aggregated global word-word co-occurrence statistics from a corpus, and the resulting representations showcase interesting linear substructures of the word vector space. The GloVe model is trained on the non-zero entries of a global word-word co-occurrence matrix, which tabulates how frequently words co-occur with one another in a given corpus. Populating this matrix requires a single pass through the entire corpus to collect the statistics\cite{pennington:14}.

By using word2vec and glove algorithms, we produced word vectors separately for the words in the reviews. That is, we have two-word vector sets for every review dataset. We obtained the best results with our methods when the lengths of word2vec and glove vectors were 300. We eliminated stop-words and did not use word vectors of stop-words. We used these vectors within our weighted averaged review vector (WARV) and convolutional neural network(CNN) architecture described in Kim \cite{Kim:14}. In the following sections, we will explain the methods in detail.

\subsection*{Averaged Review Vector (ARV) and Weighted Averaged Review Vector (WARV)}

Since word vectors represent semantic similarity between words, if we can produce a vector that represents semantic similarity for a review or sentence from word vectors, then it will be easy to learn the semantic polarity of the reviews or sentences. For that purpose, if we find a mean vector of a review from word vectors of that review, then we have a new vector representing the review, which will be semantically similar to words in the review. Therefore, we created review vectors from normalized word vectors of words in the review by averaging the word vectors. In this case, all word vectors of the words in a review will be an input of our method, and the output will be a review vector that has the same dimension as word vectors. For every review, we found averaged vectors of all reviews. We produced word vectors by using word2vec and glove algorithms. Then we created averaged review vectors (ARV) from word2vec and glove vectors separately, and we concatenated them. That is, we produced new combined word vectors from  word2vec vectors with size $M$ and glove vectors with size $M$. Obviously, the dimension of combined word vectors is $2M$. For any review contains $N$ words, we have averaged $N$ word vectors with size $M$: $[\textbf{wv}_{1}, \textbf{wv}_{2},...,\textbf{wv}_{N}]$. Mathematically, we found normalized vectors of the word vectors by
\begin{equation}
\hat{\textbf{wv}}_{i} = \frac{\textbf{wv}_{i}}{|\textbf{wv}_{i}|}
\end{equation}
where
\begin{equation}
|\textbf{wv}_{i}| = \sqrt{wv^{2}_{i1} + wv^{2}_{i2} + ... wv^{2}_{iM}}
\end{equation}

Then, we found the weighted averaged vectors of normalized word vectors of a review from word2vec vectors and glove vectors separately and then concatenated them. For that purpose, we used the following equation, which finds the weighted average of the $N$ word vectors. Note that for $b_{i} = 1$ for every $i$ we will obtain averaged review vectors.

\begin{equation}
\textbf{warv} = \frac{b_{i} * \textbf{wv}_{i} }{N}
\end{equation}
where $b_{i}$ represents weights for word vectors. Note that for $b_{i} = 1$ for every $i$ we will obtain averaged review vectors. We used different weights and we obtained the best result with the below weight for the datasets.
\begin{equation}
b_{i} = \frac{P(w_{i} \ge 0)}{P(w_{i} \le 0)}
\end{equation}
where we represented the probability of the positive polarity of the word i as $P(w_{i} \ge 0)$ and the probability of the negative polarity of the word i as $P(w_{i} \le 0)$. Obviously,
\begin{equation}
P(w_{i} \ge 0) + P(w_{i} \le 0) = 1
\end{equation}

In this way, we obtain the averaged vectors as shown in Table\ref{review-matrix} for every review by using Word2vec and Glove vectors of the reviews as shown in Table\ref{review-word-matrix}. Hence, for every review, we have $2M$ features to learn. By using Feed-forward Neural Networks, we learned the sentiment of the review with these weighted averaged review vectors. In the next session, we shared our results for different datasets. We used the Keras framework on Tensorflow to run our feed-forward neural network model which is shown in Fig \ref{fig:nn_new}. We ran our model with different numbers of hidden layers and nodes, different activation, optimizers, and loss functions; however, we obtained the best results with the following parameters: Our model includes 3 hidden layers. The first two layers have $2M$ nodes. We used the rectifier (max value of input nodes) "RELU" as an activation function. At the last layer, we used the "sigmoid" function. Our optimizer is "Adadelta," and the loss function is "binary cross-entropy." 

\begin{table}
	\begin{center}
		\scalebox{0.6}{
			\begin{tabular}{|l|l|l|l|l|l|l|l|}
				\hline
				\bf Review ID & \bf Dimension 1 & \bf Dimension 2 & \bf Dimension 3 & \bf Dimension 4 & \bf Dimension 5 & \bf ...\\ \hline 
				\hspace{0.1cm} 0 & -0.773403 & 0.080932 & 0.285580 & 0.016509 & 0.379708 & ...\\ \hline
				\hspace{0.1cm} 1 & 0.420276 & 0.270612 & -0.784823 & -0.252692 & -0.089924 & ...\\ \hline
				\hspace{0.1cm} 2 & -0.312554 & -0.435228 & -0.589917 & 0.268841 & 0.682278 & ...\\ \hline
				\hspace{0.1cm} 3 & -0.422671 & -0.821369 & -0.820449 & 0.195634 & -0.238884 & ...\\ \hline
				\hspace{0.1cm} 4 & -0.154693 & 0.000538 & -1.029475 & 0.405317 & -0.271666 & ...\\ \hline
				\hspace{0.1cm} 5 & -0.646499 & -0.795613 & -0.502919 & -1.153179 & -0.647323 & ...\\ \hline
		\end{tabular}}
	\end{center}
	\caption{\label{review-matrix} Reviews with ids (M=6)}
\end{table}

    \tikzset{%
		every neuron/.style={
			circle,
			draw,
			minimum size=1cm
		},
		neuron missing/.style={
			draw=none, 
			scale=4,
			text height=0.333cm,
			execute at begin node=\color{black}$\vdots$
		},
	}

    \begin{figure}
        \begin{tikzpicture}[x=2cm, y=1.5cm, >=stealth, samples=501]
    	\foreach \m/\l [count=\y] in {1,2,3,missing,4}
    	\node [every neuron/.try, neuron \m/.try] (input-\m) at (0,2.5-\y) {};
    	\foreach \m [count=\y] in {1,missing,2}
    	\node [every neuron/.try, neuron \m/.try ] (hidden-\m) at (2,2-\y*1.25) {};
    	\foreach \m [count=\y] in {1}
    	\node [every neuron/.try, neuron \m/.try ] (output-\m) at (4,1.5-\y*2) {};
    	\foreach \l [count=\i] in {1,2,3,d}
    	\draw [<-] (input-\i) -- ++(-1,0)
    	node [above, midway] {$Input_\l$};
    	\foreach \l [count=\i] in {1}
    	\node [above] at (hidden-\i.north) {$Hidden_\l$};
    	\node [above] at (hidden-2.north) {$Hidden_{d/2}$};
    	\foreach \l [count=\i] in {1}
    	\draw [->] (output-\i) -- ++(1,0)
    	node [above, midway] {$Output$};
    	\foreach \i in {1,...,4}
    	\foreach \j in {1,...,2}
    	\draw [->] (input-\i) -- (hidden-\j);
    	\foreach \i in {1,...,2}
    	\foreach \j in {1}
    	\draw [->] (hidden-\i) -- (output-\j);
    	\foreach \l [count=\x from 0] in {Input, Hidden, Output}
    	\node [align=center, above] at (\x*2,2) {\l \\ layer};
        \end{tikzpicture}
        \caption{\bf Our feed-forward neural network model.}
        \label{fig:nn_new}
    \end{figure}

\subsection*{Convolutional Neural Network (CNN)}
We used Convolutional Neural Networks (Kim) to learn the sentiment of the reviews by using word2vec vectors and glove vectors. We used word vectors as features directly. Our learning system is based on the Deep Convolutional Neural Network (CNN) architecture described in Kim \cite{Kim:14}. The architecture we use is shown in Fig \ref{fig:cnn}.

A review matrix is built for each input review, where each row is a vector representation of the word in the review. The review length is fixed to the maximum review length of the dataset so that all review matrices have the same dimensions. (Shorter reviews are padded with row vectors of 0s accordingly.) Each row vector of the review matrix is made up of columns corresponding to word2vec and glove vectors concatenated together.

Let $x_{i} \in R_{k}$ be the k-dimensional word vector corresponding to the i-th word in the review. A review of length n (padded where necessary) is represented as \begin{equation}
\label{1}
x_{1:n} = x_{1}  \oplus x_{2} \oplus. . .\oplus x_{n}
\end{equation}
where $\oplus$ is the concatenation operator. In general, let $x_{i:i+j}$ refer to the concatenation of words $x_{i},x_{i+1}, . . . ,x_{i+j}$. A convolution operation involves a filter $w \in R^{hk}$, which is applied to a window of h words to produce a new feature. For example, a feature $c_{i}$ is generated from a window of words $x_{i:i+h-1}$ by
\begin{equation}
\label{2}
c_{i} = f(w * x_{i:i+h-1} + b)
\end{equation}

Here $b \in R$ is a bias term and f is a non-linear function such as the hyperbolic tangent. This filter is applied to each possible window of words in the review ${x_{1:h},x_{2:h+1},...,x_{n-h+1:n}}$ to produce a feature map 
\begin{equation}
\label{3}
\textbf{c} = [c_{1}, c_{2}, . . . , c_{n-h+1}]
\end{equation}
with $c \in R^{n-h+1}$. We then apply a max-overtime pooling operation (Collobert et al. \cite{Collobert:11}) over the feature map and take the maximum value $\hat{c} = \max{\textbf{c}}$ as the feature corresponding to this particular filter. The idea is to capture the most important feature with the highest value for each feature map. This pooling scheme naturally deals with variable review lengths. One feature is extracted from one filter. The model uses multiple filters (with varying window sizes) to obtain multiple features. These features form the penultimate layer and are passed to a fully connected softmax layer whose output is the probability distribution over labels. 

For every review, we produced review matrices whose rows represent word vectors, which are obtained by concatenation of Word2vec and Glove as shown in Table \ref{review-word-matrix}. If one word does not exist in Word2vec/Glove vectors, then we fill into cells corresponding to Word2vec/Glove columns with zeros. Also, the number of rows of the review matrices is fixed to a number of words of maximum length review (N). For the reviews whose number of words is less than N, empty rows are filled by zeros. The length of our word2vec and glove vectors is 300; hence, we have an input vector whose length is 600 ($=300 + 300$).

We used Keras/Tensorflow \cite{Keras} framework to run CNN. In this framework, by default, the filters \textbf{W} are initialized randomly using the glorot uniform method, which draws values from a uniform distribution with positive and negative bounds described as in \ref{glorot}.

\begin{equation}
\label{glorot}
\textbf{W} \sim U(\frac{6}{(n_{in}+n_{out})}, \frac{-6}{(n_{in}+n_{out})})
\end{equation}

where $n_{in}$ is the number of units that feed into this unit, and $n_{out}$ is the number of units this result is fed to. Again, we used 600 filters, which is equal to the dimension of the input.

These filters are applied at each layer of the network. That is, a discrete convolution is performed for each filter on each input data, and the results of these convolutions are fed to the next layer of convolutions or a fully connected layer.

During training, the values in the filters are optimized with backpropagation by using a loss function. We used the binary cross entropy for positive/negative sentiments (IMDB dataset\cite{Stanford:11}) and categorical cross-entropy loss for positive/negative/neutral sentiments (Semeval 2016 dataset\cite{Semeval5:16}).  

    \begin{table}
	\begin{center}
		\scalebox{0.6}{
			\begin{tabular}{|l|l|l|l|l|l|l|l|}
				\hline
				\bf No & \bf Word & \bf Word2Vec 1 & \bf ... & \bf Word2Vec 300 & \bf Glove 1 & \bf ... & \bf Glove 300\\ \hline 
				1 & storm-lashed & 0.536650 & ... & -0.047624 & 0.024453 & ... & 0.418517 \\ \hline 
				2 & lended & 0.182570 & ... & 0.119859 & 0.203666 & ... & 0.408951 \\ \hline
				3 & maries & 0.309272 & ... & -0.293730 & 0.374721 & ... & -0.165671 \\ \hline 
				4 & sinfully & 0.304134 & ... & -0.236288 & 0 & ... & 0 \\ \hline
				5 & weekend & 0.436948 & ... & 0.453432 & 0.038075 & ... & -0.087220 \\ \hline
				6 & bernie's' & 0.269693 & ... & 0.545252 & 0.094293 & ... & -0.117270 \\ \hline
				7 & chique & -0.238754 & ... & 0.708557 & 0.062137 & ... & 0.267371 \\ \hline
				8 & spiritualistic & 0.211577 & ... & 0.226885 & 0.304304 & ... & 0.006760 \\ \hline 
				9 & half-awake & 0 & ... & 0 & -0.093967 & ... & 0.564555 \\ \hline
				10 & eddi & -0.260920 & ... & 0.248055 & -0.065446 & ... & -0.259685 \\ \hline
				... & ... & 0 & ... & 0 & 0 & ... & 0 \\ \hline
				N & ... & 0 & ... & 0 & 0 & ... & 0 \\ \hline
			\end{tabular}}
		\end{center}
		\caption{\label{review-word-matrix} IMDB Reviews with ids}
    \end{table}

    \tikzset{ 
	table/.style={
	  matrix of nodes,
	  row sep=-\pgflinewidth,
	  column sep=-\pgflinewidth,
	  nodes={rectangle,draw=black,text width=1.25ex,align=center},
	  text depth=0.25ex,
	  text height=1ex,
	  nodes in empty cells
	  },
	texto/.style={font=\footnotesize\sffamily},
	title/.style={font=\small\sffamily}
	}	
	\newcommand\CellText[2]{%
	  \node[texto,left=of mat#1,anchor=west]
	  at (mat#1.west)
	  {#2};
	}
	\newcommand\SlText[2]{%
	  \node[texto,above=0.75cm of mat#1,anchor=west,rotate=0]
	  at ([xshift=-2ex]mat#1.north)
	  {#2};
	}
	\newcommand\SlTextShort[2]{%
		\node[texto,above=of mat#1,anchor=west,rotate=0]
		at ([xshift=-1ex]mat#1.north)
		{#2};
	}
	\newcommand\RowTitle[2]{%
	\node[title,left=2.1cm of mat#1,anchor=west]
	  at (mat#1.north west)
	  {#2};
	}
	\newcommand\RowTitlePer[2]{%
		\node[title,left=of mat#1,anchor=west,rotate=90]
		at (mat#1.south west)
		{#2};
	}
	\newcommand\CellTextRight[2]{%
		\node[title,right=0.5cm of mat#1,anchor=west]
		at (mat#1.west)
		{#2};
	}
    \begin{figure}[!h]
        \centering
	\begin{tikzpicture}[node distance =10pt and 2cm]
	\matrix[rounded corners,table] (mat11) 
	{
		|[fill=red]| & |[fill=red]| & |[fill=red]| & |[fill=red]|\\
		|[fill=red]| & |[fill=red]| & |[fill=red]| & |[fill=red]|\\
		 & & & \\
		 & & & \\
		|[fill=yellow]| & |[fill=yellow]| & |[fill=yellow]| & |[fill=yellow]|\\
		|[fill=yellow]| & |[fill=yellow]| & |[fill=yellow]| & |[fill=yellow]|\\
	};
	\matrix[rounded corners, table,right=of mat11] (mat12) 
	{
		|[fill=red]| \\
		 \\
		 \\
		 \\
		|[fill=yellow]| \\
	};
	\matrix[rounded corners, table,below=of mat11] (mat21) 
	{
		|[fill=blue]| & |[fill=blue]| & |[fill=blue]| & |[fill=blue]|\\
		|[fill=blue]| & |[fill=blue]| & |[fill=blue]| & |[fill=blue]|\\
		& & & \\
		& & & \\
		|[fill=orange]| & |[fill=orange]| & |[fill=orange]| & |[fill=orange]|\\
		|[fill=orange]| & |[fill=orange]| & |[fill=orange]| & |[fill=orange]|\\
	};
	\matrix[rounded corners, table,right=of mat21] (mat22) 
	{
		|[fill=blue]| \\
		\\
		\\
		\\
		|[fill=orange]| \\
	};
	\matrix[rounded corners, table,below=1cm of mat21] (mat31) 
	{
		|[fill=green]| & |[fill=green]| & |[fill=green]| & |[fill=green]|\\
		|[fill=green]| & |[fill=green]| & |[fill=green]| & |[fill=green]|\\
		& & & \\
		& & & \\
		|[fill=brown]| & |[fill=brown]| & |[fill=brown]| & |[fill=brown]|\\
		|[fill=brown]| & |[fill=brown]| & |[fill=brown]| & |[fill=brown]|\\
	};
	\matrix[rounded corners, table,right=of mat31] (mat32) 
	{
		|[fill=green]| \\
		\\
		\\
		\\
		|[fill=brown]| \\
	};
	\matrix[rounded corners, table,right=2cm of mat22] (mat23) 
	{
		|[fill=yellow]| \\
		|[fill=blue]| \\
		\\
		|[fill=green]| \\
	};
	\matrix[rounded corners, table,right=of mat23] (mat24) 
	{
		|[fill=gray]| \\
		|[fill=gray]| \\
		|[fill=gray]| \\
	};
	
	\SlTextShort{11-1-1}{1}
	\SlTextShort{11-1-2}{2}
	\SlTextShort{11-1-3}{...}
	\SlTextShort{11-1-4}{d1+d2}
	\SlText{11-1-1}{Word vectors}
	\SlText{12-1-1}{Features after filters}
	\SlText{23-1-1}{Max-Pooling}
	\SlText{24-1-1}{Softmax output}
	\SlText{31-1-1}{\textbf{..........}}
	\RowTitle{11}{Review 1};
	\CellText{11-1-1}{Word 1};
	\CellText{11-2-1}{Word 2};
	\CellText{11-3-1}{Word 3};
	\CellText{11-4-1}{......};
	\CellText{11-6-1}{Word N};
	\RowTitle{21}{Review 2};
	\CellText{21-1-1}{Word 1};
	\CellText{21-2-1}{Word 2};
	\CellText{21-3-1}{Word 3};
	\CellText{21-4-1}{......};
	\CellText{21-6-1}{Word N};
	\RowTitle{31}{Review N};
	\CellText{31-1-1}{Word 1};
	\CellText{31-2-1}{Word 2};
	\CellText{31-3-1}{Word 3};
	\CellText{31-4-1}{......};
	\CellText{31-6-1}{Word N};
	\CellTextRight{24-1-1}{Positive};
	\CellTextRight{24-2-1}{Negative};
	\CellTextRight{24-3-1}{Neutral};
	\draw[->,red,line width=0.5mm] (mat11-1-4.east) -- (mat12-1-1.west);
	\draw[->,red,line width=0.5mm] (mat11-2-4.east) -- (mat12-1-1.west);
	\draw[->,yellow,line width=0.5mm] (mat11-5-4.east) -- (mat12-5-1.west);
	\draw[->,yellow,line width=0.5mm] (mat11-6-4.east) -- (mat12-5-1.west);
	\draw[->,blue,line width=0.5mm] (mat21-1-4.east) -- (mat22-1-1.west);
	\draw[->,blue,line width=0.5mm] (mat21-2-4.east) -- (mat22-1-1.west);
	\draw[->,orange,line width=0.5mm] (mat21-5-4.east) -- (mat22-5-1.west);
	\draw[->,orange,line width=0.5mm] (mat21-6-4.east) -- (mat22-5-1.west);
	\draw[->,green,line width=0.5mm] (mat31-1-4.east) -- (mat32-1-1.west);
	\draw[->,green,line width=0.5mm] (mat31-2-4.east) -- (mat32-1-1.west);
	\draw[->,brown,line width=0.5mm] (mat31-5-4.east) -- (mat32-5-1.west);
	\draw[->,brown,line width=0.5mm] (mat31-6-4.east) -- (mat32-5-1.west);
	\draw[->,red,line width=0.5mm] (mat12-1-1.east) -- (mat23-1-1.west);
	\draw[->,yellow,line width=0.5mm] (mat12-5-1.east) -- (mat23-1-1.west);
	\draw[->,blue,line width=0.5mm] (mat22-1-1.east) -- (mat23-2-1.west);
	\draw[->,orange,line width=0.5mm] (mat22-5-1.east) -- (mat23-2-1.west);
	\draw[->,green,line width=0.5mm] (mat32-1-1.east) -- (mat23-4-1.west);
	\draw[->,brown,line width=0.5mm] (mat32-5-1.east) -- (mat23-4-1.west);
	\draw[->,gray,line width=0.5mm] (mat23-1-1.east) -- (mat24-1-1.west);
	\draw[->,gray,line width=0.5mm] (mat23-4-1.east) -- (mat24-3-1.west);
	\end{tikzpicture}
        \caption{\bf CNN Sentence Classification.}
        \label{fig:cnn}
    \end{figure}

\subsection*{Ensemble}

After learning by using training data, we combine the results of different learning algorithms by using validation data sets. We used two different ensemble approaches: As a first ensemble approach (Ensemble-1), we used the log probability scores approach in Mesnil's study\cite{Mesnil:15}.

As the second one, we used a neural network over validation sets (Ensemble-2) at Fig \ref{fig:ensemble}. In this case, by using validation data sets, our learning algorithms produce results. The results contain probabilities for each class. Then we used these results as features of the ensemble learning algorithm. For that purpose, we used different learning algorithms to ensemble the methods, and we obtained the best accuracies with logistic regression and neural networks. Our neural network model contains three hidden layers whose number of nodes is equal to the input length. We used Rectified Linear Unit as the activation function of the layers, Sigmoid function to estimate the class or label, AdaDelta\cite{Zeiler:12} as an optimizer, and cross-entropy as a loss function.  Then we tested our ensemble methods with test datasets again. Our ensemble method produced better results than every single learning algorithm.

\definecolor{redi}{RGB}{255,38,0}
\definecolor{redii}{RGB}{200,50,30}
\definecolor{rediii}{RGB}{150,50,30}
\definecolor{yellowi}{RGB}{255,251,0}
\definecolor{bluei}{RGB}{0,150,255}
\definecolor{blueii}{RGB}{135,247,210}
\definecolor{blueiii}{RGB}{91,205,250}
\definecolor{blueiv}{RGB}{115,244,253}
\definecolor{bluev}{RGB}{1,58,215}
\definecolor{orangei}{RGB}{240,143,50}
\definecolor{yellowii}{RGB}{222,247,100}
\definecolor{greeni}{RGB}{166,247,166}
\definecolor{greenii}{RGB}{166,200,166}
\tikzset{ 
    table/.style={
      matrix of nodes,
      row sep=-\pgflinewidth,
      column sep=-\pgflinewidth,
      nodes={rectangle,draw=black,text width=1.25ex,align=center},
      text depth=0.25ex,
      text height=1ex,
      nodes in empty cells
      },
    texto/.style={font=\footnotesize\sffamily},
    title/.style={font=\small\sffamily}
}
\tikzset{vertex/.style = {shape = circle,
    text           = black,
    inner sep      = 2pt,
    outer sep      = 0pt,
    minimum size   = 55 pt}
}
\tikzset{edge/.style   = {thick,
    double          = orange,
    double distance = 1pt}
}
    
\begin{figure}
    \centering
    \begin{tikzpicture}[node distance =10pt and 0.5cm]
        \matrix[table] (mat11) 
        {
        	&  &  & \\
        	&  &  & \\
        	&  &  & \\
        	&  &  & \\
        	&  &  & \\
        };
        \node[vertex, right=of mat11](mat12){WARV};
        \matrix[table,right=of mat12] (mat13) 
        {
        	|[fill=redii]| & |[fill=redii]| & |[fill=redii]| \\
        	|[fill=redii]| & |[fill=redii]| & |[fill=redii]| \\
        	|[fill=redii]| & |[fill=redii]| & |[fill=redii]| \\
        	|[fill=redii]| & |[fill=redii]| & |[fill=redii]| \\
        	|[fill=redii]| & |[fill=redii]| & |[fill=redii]| \\
        };
        \matrix[table,below=of mat11] (mat21) 
        {
        	&  &  & \\
        	&  &  & \\
        	&  &  & \\
        	&  &  & \\
        	&  &  & \\
        };
        \node[vertex, right=of mat21](mat22){ARV};
        \matrix[table,below=of mat13] (mat23) 
        {
        	|[fill=bluei]| & |[fill=bluei]| & |[fill=bluei]| \\
        	|[fill=bluei]| & |[fill=bluei]| & |[fill=bluei]| \\
        	|[fill=bluei]| & |[fill=bluei]| & |[fill=bluei]| \\
        	|[fill=bluei]| & |[fill=bluei]| & |[fill=bluei]| \\
        	|[fill=bluei]| & |[fill=bluei]| & |[fill=bluei]| \\
        };
        \matrix[table,below=of mat21] (mat31) 
        {
        	&  &  & \\
        	&  &  & \\
        	&  &  & \\
        	&  &  & \\
        	&  &  & \\
        };
        \node[vertex,right=of mat31](mat32){CNN};
        \matrix[table,below=of mat23] (mat33) 
        {
        	|[fill=greeni]| & |[fill=greeni]| & |[fill=greeni]| \\
        	|[fill=greeni]| & |[fill=greeni]| & |[fill=greeni]| \\
        	|[fill=greeni]| & |[fill=greeni]| & |[fill=greeni]| \\
        	|[fill=greeni]| & |[fill=greeni]| & |[fill=greeni]| \\
        	|[fill=greeni]| & |[fill=greeni]| & |[fill=greeni]| \\
        };
        \node[vertex,right=of mat23](mat24){Ensemble};
        \matrix[table,right=of mat24] (mat25) 
        {
        	|[fill=yellowi]| & |[fill=yellowi]| & |[fill=yellowi]|\\
        	|[fill=yellowi]| & |[fill=yellowi]| & |[fill=yellowi]|\\
        	|[fill=yellowi]| & |[fill=yellowi]| & |[fill=yellowi]|\\
        	|[fill=yellowi]| & |[fill=yellowi]| & |[fill=yellowi]|\\
        	|[fill=yellowi]| & |[fill=yellowi]| & |[fill=yellowi]|\\
        };
        \SlText{11}{Input Data}
        \SlText{12}{Methods}
        \SlText{13}{Output Probabilities}
        \SlText{25}{Output}
        \draw[->,black] (mat11-3-4.east) -- (mat12.west);
        \draw[->,black] (mat21-3-4.east) -- (mat22.west);
        \draw[->,black] (mat31-3-4.east) -- (mat32.west);
        \draw[->,black] (mat12.east) -- (mat13-3-1.west);
        \draw[->,black] (mat22.east) -- (mat23-3-1.west);
        \draw[->,black] (mat32.east) -- (mat33-3-1.west);
        \draw[->,black] (mat13-3-3.east) -- (mat24.west);
        \draw[->,black] (mat23-3-3.east) -- (mat24.west);
        \draw[->,black] (mat33-3-3.east) -- (mat24.west);
        \draw[->,black] (mat24.east) -- (mat25-3-1.west);
    \end{tikzpicture}
    \caption{\bf Ensemble of Learning Algorithms.}
    \label{fig:ensemble}
\end{figure}

\section*{Experiments}
In this work, we test our models and other existing models with different datasets. We used Stanford IMDB Reviews\cite{Stanford:11}, SEMEVAL 2016 Task 5 dataset\cite{Semeval5:16} and YELP dataset\cite{Yelp:18}. In this section, we will explain experiments and results.

\subsection*{IMDB Reviews}
Stanford IMDB dataset of reviews contains 100,000 movie reviews in English. 25,000 of these reviews are labeled as positive, the other 25,000 reviews are labeled as negative and the remaining 50,000 reviews are unlabeled. In fact, IMDB reviews are labeled numbers that are between 1 and 10. However, this dataset contains only reviews with 1 to 4 as negative and reviews with 7 to 10 as positive. Therefore, there are only two labels: positive and negative.

For IMDB reviews, we produced word2vec and glove vectors from IMDB reviews. The length of the vectors is 300, and their window sizes are 5. We run our methods by using word2vec vectors(\_wv) and glove vectors(\_gl) separately and combined(\_wvgl). Therefore, the number of features of our Averaged Review Vectors (ARV) and Weighted Averaged Review Vectors (WARV) is 300 for separate runs and 600 for combined runs. In the same manner, for our CNN model, the number of features is 300 (separate run) and 600 (combined run).

    \begin{table}
	\centering
	\caption{Accuracies of Methods Before Ensemble for IMDB dataset}
	\label{imdb_single_acc}
	\scalebox{0.8}{
	\begin{tabular}{|l|c|l|}
		\hline
		\textbf{Methods} & \textbf{Accuracies} & \textbf{Explanation}\\ \hline
		RNNLM 		& 86.30 & Recurrent Neural Network (Mesnil et al.\cite{Mesnil:15}) \\ \hline
		PV 			& 88.45 & Paragraph Vectors (Mesnil et al.\cite{Mesnil:15}) \\ \hline
		NBSVM 		& 91.87 & NB-SVM TriGram (Wang et al.\cite{Mesnil:15}\cite{Wang:12}) \\ \hline
		ARV\_wv 	& 88.05 & Average Vectors of Reviews with Neural Network (Word2Vec)\\ \hline
		ARV\_gl 	& 84.54 & Average Vectors of Reviews with Neural Network (Glove)\\ \hline
		ARV 		& 88.07 & Average Vectors of Reviews with Neural Network (Word2Vec + Glove)\\ \hline
		WARV\_wv 	& 93.91 & Weighted Average Vectors of Reviews with Neural Network (Word2Vec) \\ \hline
		WARV\_gl 	& 86.24 & Weighted Average Vectors of Reviews with Neural Network (Glove) \\ \hline
		WARV 		& \bf 94.286 & Weighted Average Vectors of Reviews with Neural Network (Word2Vec + Glove)\\ \hline
		CNN 		& 89.19 & Convolutional Neural Network with Word Vectors \\ \hline
	\end{tabular}}
    \end{table}

As seen in Table \ref{imdb_single_acc}, with \textbf{\% 94.286} accuracy, our WARV method has better accuracy values than N-Gram, RNNLM, and Mesnil's paragraph vector. Also, our WARV method is computationally more efficient than Mesnil's paragraph vectors (PV) (Mesnil et al.\cite{Mesnil:15}).

    \begin{table}
	\centering
	\caption{Accuracies of Methods After Ensemble for IMDB dataset}
	\label{imdb_ensemble_acc}
	\scalebox{0.8}{
	\begin{tabular}{|l|c|c|}
		\hline
		\textbf{Methods}	&	\textbf{Ensemble-1}	&	\textbf{Ensemble-2} \\ \hline
		\bf RNNLM, PARAGRAPH (Mesnil's) 	&	90.14	&	89.58  \\ \hline 	
		\bf RNNLM, NBSVM (Mesnil's)	 	&	92.08	&	91.764 \\ \hline			
		PARAGRAPH, NBSVM 	&	92.13	&	92.34  \\ \hline				
		WARV, RNNLM 	 	&	94.64	&	94.412 \\ \hline		
		WARV, PARAGRAPH 	&	94.44	&	94.488 \\ \hline 	 	
		\bf WARV, NBSVM 	 	&	\bf 95.02	&	0.94688 \\ \hline		
		WARV, ARV 	 		&	94.57	&	94.34  \\ \hline		
		WARV, CNN 	 		&	94.65	&	94.488 \\ \hline	
		\hline
		\bf RNNLM, PARAGRAPH, NBSVM (Mesnil's)	 	& \bf 92.37 & 92.352	\\ \hline	 
		WARV, PARAGRAPH, NBSVM 			& 94.33 & 94.768	\\ \hline	 
		WARV, PARAGRAPH, ARV 			& 92.74 & 94.464	\\ \hline	 
		WARV, PARAGRAPH, CNN 			& 93.83 & 94.672	\\ \hline	 
		WARV, NBSVM, ARV 	 			& 94.13 & 94.684	\\ \hline	 
		\bf WARV, NBSVM, CNN 	 			& 94.47 & \bf 95.032	\\ \hline	 
		WARV, ARV, CNN 	 				& 93.74 & 94.56		\\ \hline	 
		RNNLM, PARAGRAPH, WARV 	 		& 94.44 & 94.544	\\ \hline	 
		RNNLM, WARV, ARV 	 			& 93.98 & 94.46		\\ \hline	 
		RNNLM, WARV, CNN 	 			& 94.82 & 94.576	\\ \hline	 
		RNNLM, NBSVM, WARV 	 			& 94.89 & 94.696	\\ \hline	 
		\hline
		RNNLM, PARAGRAPH, NBSVM, WARV 	& 94.41 & 94.768	\\ \hline
		RNNLM, PARAGRAPH, ARV, WARV 	& 93.18 & 94.552	\\ \hline
		PARAGRAPH, NBSVM, ARV, WARV 	& 93.74 & 94.768	\\ \hline
		\bf PARAGRAPH, NBSVM, CNN, WARV 	& 94.10 & \bf 95.028	\\ \hline
		PARAGRAPH, WARV, ARV, CNN 	 	& 93.35 & 94.668	\\ \hline
		RNNLM, NBSVM, ARV, WARV 	 	& 94.33 & 94.70		\\ \hline
		\bf RNNLM, NBSVM, CNN, WARV 	 	& 94.56 & \bf 95.032	\\ \hline
		\bf NBSVM, WARV, ARV, CNN 	 		& 94.05 & \bf 95.032	\\ \hline
		RNNLM, WARV, ARV, CNN 	 		& 93.99 & 94.64		\\ \hline
		PARAGRAPH, WARV, ARV, CNN 	 	& 93.35 & 94.668	\\ \hline
		\hline
		WARV, PARAGRAPH, NBSVM, ARV, CNN 	& 93.69	& 95.028	\\ \hline
		\bf RNNLM, WARV, NBSVM, ARV, CNN 	 	& 94.17	& \bf 95.032	\\ \hline
		RNNLM, PARAGRAPH, WARV, ARV, CNN 	& 93.49	& 94.732	\\ \hline
		\bf RNNLM, PARAGRAPH, NBSVM, WARV, CNN 	& 94.14	& \bf 95.028	\\ \hline
		RNNLM, PARAGRAPH, NBSVM, ARV, WARV 	& 93.87	& 94.768	\\ \hline
		\hline
		\bf RNNLM, PARAGRAPH, NBSVM, ARV, CNN, WARV	& 93.86 & \bf 95.028	\\ \hline
	\end{tabular}}
    \end{table}

When our weighted averaged review vectors (WARV) are ensembled with NBSVM, they reach an accuracy value \textbf{\% 95.02} as shown in Table \ref{imdb_ensemble_acc}. When we ensemble three methods, the best accuracy (\textbf{\% 95.032}) is obtained by the ensemble of WARV, NBSVM, and CNN as shown in Table \ref{imdb_ensemble_acc}. Also, these two results are better than Mesnil's results.
				
When we ensemble four methods or five methods, the best accuracy is again \textbf{\% 95.032} as shown in Table \ref{imdb_ensemble_acc}. When we ensemble all six methods, accuracy becomes \textbf{\% 95.028}. These results are nearly close to other results that were obtained by using Bidirectional Transformers BERT embeddings \cite{Bert} in some studies \cite{Haonan:19}\cite{Wang:21}\cite{Chi:19}\cite{Quizhe:19} that reached \% 96 accuracy.

Note that we didn't share the ensemble results of the combinations whose accuracies are below \% 94 except for ensembles in Mesnil's study\cite{Mesnil:15}.

\subsection*{Semeval 2016 Task 5 dataset}
In English, two domain-specific datasets for consumer electronics (laptops) and restaurants, consisting of over 1000 review texts (approx. 6K sentences) with fine-grained human annotations (opinion target expressions, aspect categories, and polarities) will be provided for training/development. In particular, the SE-ABSA15 train and test datasets for restaurants and laptops (with some corrections) will be made available as training data. They consist of 800 review texts (4500 sentences) annotated with approx. 15000 unique label assignments (Entity, Aspect, polarity). The laptop dataset consists of 450 review texts (2500 sentences) annotated with 2923 {Entity\#Aspect, polarity} tuples. The restaurant dataset consists of 350 review texts (2000 sentences) annotated with 2499 {Entity\#Aspect, polarity} tuples. All datasets will be enriched with text-level annotations. Also, datasets exist for other languages rather than English. These languages are Arabic, Chinese, Dutch, French, Russian, Spanish, Turkish\cite{Semeval5:16}.

We used our two methods on the English restaurant dataset. We produced word2vec and glove vectors from the Yelp dataset. For that purpose, firstly, we get only restaurant-related reviews of the Yelp dataset, then we create word vectors. After that, we applied our two methods (ARV and CNN) to the Semeval 2016 Task 5 dataset to learn polarity. Especially our ARV method produced high accuracy results (\textbf{$87.7\%$}). It is very close to the best result of Semeval 2016 Task 5 as shown in Table \ref{semeval_acc} and the study \cite{Reddy:20} (\textbf{$87.8\%$}) with BERT embedding \cite{Bert}. 

We ensembled our two results by using a neural network. For this purpose, we used probability outputs of ARV and CNN. Since we have three classes (positive, neutral, and negative), every method has 3 probability values. Therefore, we used 6 probability values of two methods as our learning features. Since Semeval 2016 dataset is very small, we have no opportunity to use some parts of the dataset for validation. Instead of probability values on validation data, we used probabilities of the training dataset. That is, our ensemble feed-forward neural network used probabilities that were produced over the training dataset of Semeval 2016 Task5 by our ARV and CNN methods. Then, we used the test dataset of Semeval 2016 Task 5 to evaluate our ensemble model. As shown in Table \ref{semeval_acc}, our ensemble produced results that are very close to the state-of-the-art value (\textbf{$88.242\%$}) for the Semeval 2016 Task 5 dataset. We used the Keras framework to run our ensemble neural network model. Our ensemble learning neural network includes 3 hidden layers with 600 nodes, which is equal to the number of features. As an activation function, we used "scaled exponential linear units" (SELUs), which induce self-normalizing properties\cite{Klambauer:17}. At the last layer, we used the "softmax" function. Our optimizer is "Adadelta," and the loss function is "binary cross-entropy."

Note that we did not share the results of methods RNNLM, NBSVM, and PV, which were compared with our methods in the previous section for this dataset since they suffered from a small dataset. They could not learn accurately, and the results of them were very low.

Pretrained transformer-based models have recently been used in sentiment analysis \cite{Amira:23}. We also used average pretrained transformer-based models BERT vectors \cite{Bert:19} to compare our results with transformer-based models. For that purpose, we got vectors of tokens in the last layer of the BERT model, then we calculated the average for each review, which is the same method as our other models. Then, we used these vectors instead of our Word2Vec and Glove vectors to learn the sentiment of the reviews. For Semeval 2016 dataset, we obtained 86.39\% accuracy which is less than our ensemble results. Also for the IMDB movie review dataset, we obtained 90.43\% accuracy which is less than our ensemble results.

    \begin{table}
	\centering
	\caption{Accuracies of Methods Before Ensemble (Ensemble-2) for Semeval Restaurant dataset}
	\label{semeval_acc}
	\scalebox{0.8}{
		\begin{tabular}{|l|c|l|}
			\hline
			\textbf{Methods} 	& \textbf{Accuracies} & \textbf{Explanation}\\ \hline
			\bf ARV, WARV and CNN 	& \bf 88.669 	& Ensemble of ARV, WARV and CNN by Neural Network \\ \hline
			\bf WARV and CNN 	& \bf 88.630 	& Ensemble of WARV and CNN by Neural Network \\ \hline
			\bf ARV and CNN 	& \bf 88.475 	& Ensemble of ARV and CNN by Neural Network \\ \hline
			XRCE 				& 88.126 		& Semeval 2016 Task 5 Slot 1 Best Accuracy\cite{Semeval5:16} \\ \hline
			IIT-T				& 86.729		& Semeval 2016 Task 5 Slot 1 Second Best Accuracy\cite{Semeval5:16} \\ \hline
			NileT				& 85.448		& Semeval 2016 Task 5 Slot 1 Third Best Accuracy \\ \hline
			IHS-R				& 83.935		& Semeval 2016 Task 5 Slot 1 Fourth Best Accuracy\cite{Semeval5:16} \\ \hline
			ECNU				& 83.236		& Semeval 2016 Task 5 Slot 1 Fifth Best Accuracy\cite{Semeval5:16} \\ \hline
			\bf CNN 			& \bf 83.120 	& Convolutional Neural Network with Word Vectors \\ \hline
			\bf WARV 			& \bf 82.654 	& Weighted Averaged Review Vectors with Neural Network \\ \hline
			INSIG				& 82.072		& Semeval 2016 Task 5 Slot 1 Sixth Best Accuracy\cite{Semeval5:16} \\ \hline
			\bf ARV 			& \bf 81.831 	& Averaged Review Vectors with Neural Network \\ \hline 
			\hline
			\bf basel			& 76.484		& Base value of Semeval 2016 Task 5 Slot 1\cite{Semeval5:16} \\ \hline
		\end{tabular}}
    \end{table}

\subsection*{Yelp dataset}
We tested a weighted average review vector with randomly chosen restaurant-related reviews. When we used 100.000 reviews (50.000 positive, 50.000 negative) for training and 100.000 reviews (50.000 positive, 50.000 negative) for testing, our accuracy was \% 73.811. When we used 500.000 reviews (250.000 positive, 250.000 negative) for training and 500.000 reviews (250.000 positive, 250.000 negative) for testing,  our accuracy was \% 77.496. Again, we used word vectors that were produced from the Yelp dataset. This dataset \cite{Yelp:18} is a subset of businesses, reviews, and user data for use in personal, educational, and academic purposes. Available in both JSON and SQL files. Each file is composed of a single object type, one JSON object per line. According to the business IDs, we divided the review file into two separate files; one of them consists of restaurant reviews and the other one consists of non-restaurant reviews. For restaurants, we used approximately two million reviews to produce word vectors. For that purpose, we used two files from the Yelp dataset:

\begin{itemize}
\item business.json: Contains business data including location data, attributes, and categories.
\item review.json: Contains full review text data, including the user ID that wrote the review and the business ID the review is written for.
\end{itemize}

\section*{Conclusion}
In this work, we used a combination of Word2vec and Glove vectors. For the CNN model, these vectors were directly used to create review matrices. For our averaged vector model, for a review, we found averaged vectors from word vectors of the word in the review. For our weighted averaged vector model, firstly, we multiplied word vectors by predefined values such as the ratio of numbers of positive and negative words. Then, we found averaged vectors. Then, we used these vectors as input to learning algorithms. We compared our results with different methods and with different datasets. And we are very close to state-of-the-art results. For the IMDB dataset, we improved accuracy by up to three percent. For the Semeval-2016 Task-5 dataset, we improved accuracy a little bit. Our weighted averaged vector model obtained high accuracy values for the IMDB dataset and it is computationally very efficient. Building word vectors only once is enough for our method, although the paragraph vector method needs to build word vectors for every new input data. Also, our ensemble results are very high, and our ensemble method produced better results than every single learning algorithm. For our method, the next step is a lookup operation to find the word vector of the words and calculate the weighted average of the word vectors. We will continue to evaluate our models and their ensembles with different datasets in different domains. Also, our models don't include language-specific features. Hence, we will test our model in different languages rather than English. Also, we will try to find different weights that may produce better results.

\bibliographystyle{unsrt}  


\end{document}